\documentclass{article}

\usepackage[english]{babel}

\usepackage[letterpaper,top=2cm,bottom=2cm,left=3cm,right=3cm,marginparwidth=1.75cm]{geometry}

\usepackage{amsmath}
\usepackage{graphicx}
\usepackage[colorlinks=true, allcolors=blue]{hyperref}

\title{Embodied AI in Action\\Insights from SAE World Congress 2026 on Safety, Trust, Robotics, and Real-World Deployment}
\author{Jan-Mou Li, Paul Schmitt, Wei Tong, Majed Mohammed, \\Akshay Chalana, Arpan Kusari, Edward Griffor}

\begin{document}
\maketitle

\tableofcontents

\begin{abstract}
Embodied artificial intelligence is rapidly moving from research into real-world systems such as autonomous vehicles, mobile robots, and industrial machines. As these systems become more capable of perceiving, deciding, and acting in dynamic environments, they also introduce new challenges in safety, trust, governance, and operational reliability.

This white paper summarizes key insights from the SAE World Congress 2026 panel session \textit{Embodied AI in Action}, which brought together experts from automotive, robotics, artificial intelligence, and safety engineering. The discussion highlighted the need to treat embodied AI as a systems challenge requiring engineering rigor, lifecycle governance, human-centered design, and evolving standards.

The paper provides practical perspectives for executives, policymakers, and technical leaders seeking to adopt embodied AI responsibly. The panel reached broad agreement that long-term success will depend not only on advances in AI capability, but equally on safe and trustworthy deployment.

\end{abstract}

\section{Executive Summary}
Embodied artificial intelligence is rapidly moving from research environments into the physical world. Autonomous vehicles, mobile robots, warehouse systems, industrial machines, and assistive platforms are increasingly expected to perceive their surroundings, make decisions, and act safely alongside people. This transition presents major opportunities for productivity, mobility, and quality of life, but it also introduces new challenges in safety assurance, trust, governance, and system reliability.

At SAE World Congress 2026, a panel of experts from industry, consulting, startups, and academia convened to discuss the current state of embodied AI and the path toward responsible deployment. Drawing from backgrounds spanning automotive engineering, robotics, artificial intelligence, machine learning operations, safety standards, and systems integration, the panel examined what is required to transition embodied AI from promising demonstrations to dependable real-world products.

Several themes emerged consistently across the discussion. First, embodied AI should be viewed as a systems challenge rather than only a software challenge. Safe deployment depends not only on algorithms, but also on sensors, hardware robustness, operational design limits, human interaction, communications, cybersecurity, and organizational processes. Second, trust will be earned through performance that is observable, explainable, and repeatable under real operating conditions. Third, current standards and regulatory frameworks are evolving, but additional practical methods are needed to evaluate learning-enabled systems whose behavior depends on data and adaptation. Finally, progress will require collaboration across disciplines that have historically operated separately, including AI development, safety engineering, robotics, product management, and public policy.

The panel also highlighted practical examples from both automotive and robotics domains. In mobility applications, embodied AI may improve driver assistance, autonomous navigation, and traffic interaction. In robotics, embodied AI may enable machines to operate more flexibly in warehouses, healthcare environments, factories, and public spaces. In both sectors, the core challenge is similar: enabling machines to act intelligently in dynamic environments while maintaining acceptable levels of safety and reliability.

This white paper summarizes the major insights from the panel discussion, identifies common deployment barriers, and outlines practical recommendations for leaders evaluating embodied AI strategies. While the technologies continue to evolve rapidly, the panel reached broad agreement on one principle: successful adoption will depend on pairing innovation in AI capabilities with equal rigor in engineering discipline, safety thinking, and governance.

\section{What Is Embodied AI?}
Embodied artificial intelligence refers to AI systems that perceive, decide, and act through a physical device operating in the real world~\cite{duan2022survey}. Unlike software-only AI, which may generate recommendations, classify information, or automate digital workflows, embodied AI is connected to sensors, actuators, and mechanical systems that allow it to influence physical environments. In practical terms, embodied AI is what enables a machine to observe conditions around it, interpret those conditions, choose an action, and carry out that action through movement or control. While the term has been around for at least two decades~\cite{chrisley2003embodied, pfeifer2004embodied}, with the increasing use of AI across diverse domains such as robotics and autonomous vehicles, embodied AI has come to the forefront. 

Common examples include autonomous vehicles navigating traffic~\cite{bevly2016lane, stryszowski2020framework}, mobile robots moving materials through warehouses~\cite{lackner2024review, morais2025review}, robotic arms adapting to variable manufacturing tasks~\cite{bodenhagen2014adaptable, li2019survey}, drones inspecting infrastructure~\cite{seo2018drone, lattanzi2017review}, and assistive robots operating in healthcare or service environments~\cite{aymerich2023socially, holland2021service}. In each case, the AI system must function reliably amid uncertainty, changing conditions, and interactions with people.

Most embodied AI systems combine several technical functions. They often rely on perception systems to interpret sensor data, planning systems to determine possible actions, control systems to execute movement, and monitoring systems to detect faults or degraded performance. Increasingly, machine learning models are used within these functions because they can recognize patterns and adapt to complex inputs more effectively than rule-based methods alone. However, learning-enabled components also introduce challenges related to explainability, validation, and behavior outside training conditions. Furthermore, industry is moving toward end-to-end architectures. While traditional modular systems separate perception and planning, end-to-end models reason about tasks with greater computational efficiency, though they require a higher investment in post-development validation and testing due to reduced interpretability.

The technical leap to embodied AI is characterized by the implementation of a World Model \cite{ha2018recurrent}, an internal representation that allows an agent to infer the future state of the environment (e.g., predicting a cyclist’s trajectory) from minimal samples. This distinction is important because the physical world is inherently less predictable than many digital environments. Roads contain weather changes, unusual driver behavior, construction zones, and vulnerable road users. Warehouses may contain moving people, shifting inventory, unexpected obstructions, and mixed traffic between machines and workers. Public environments introduce additional complexity through crowds, variable lighting, noise, and diverse human behavior. Embodied AI systems must operate safely despite these uncertainties.

The panel discussion emphasized that embodied AI should be understood as a complete system rather than as a standalone model. A highly capable algorithm does not guarantee a safe or useful product. Outcomes depend on how software interacts with hardware, how operating limits are defined, how humans are expected to engage with the system, and how the organization manages updates, monitoring, and accountability over time.

The panel also highlighted the importance of distinguishing automation from autonomy. Traditional automation often follows predefined sequences in structured environments. Embodied AI, by contrast, is increasingly expected to make context-sensitive decisions in less structured settings. This creates new opportunities for flexibility and efficiency, but it also raises the standard for assurance and governance.

For non-technical stakeholders, a useful way to think about embodied AI is simple: it is AI that can do something physical. If it can drive, move, lift, navigate, inspect, interact, or manipulate the real world, then questions of safety, trust, and reliability become central to its adoption. That is why embodied AI has become such an important topic across mobility, robotics, manufacturing, energy, healthcare, and public policy.

\section{Why Embodied AI Matters Now}
Embodied artificial intelligence has reached an important transition point. Several forces are accelerating this transition. Improvements in machine learning, lower-cost sensors, edge computing hardware, cloud connectivity, and simulation tools have made it more practical to deploy intelligent machines outside controlled laboratory settings. At the same time, labor shortages, demographic shifts, supply chain pressures, and rising expectations for productivity are creating demand for greater automation across many industries. Organizations are seeking systems that can operate with more flexibility than traditional automation while adapting to dynamic environments. While large OEMs focus on scale, Embodied AI offers unique value to Small and Medium Enterprises (SMEs). For companies with limited staff, flexible AI-driven robots can be quickly "taught" to take over different roles, reducing the operational downtime caused by labor shortages.

This momentum also brings heightened responsibility. When AI is embodied in a vehicle, robot, or industrial machine, errors can have immediate physical consequences. A sub-optimal recommendation in software may be inconvenient, but a similar decision in the physical world can lead to safety risks resulting in property damage, disrupted operations, or erosion of public trust. As a result, deployment decisions for embodied AI require a higher standard of assurance.

Another reason embodied AI matters now is that institutions are beginning to respond. Standards bodies, regulators, insurers, and enterprise buyers are increasingly asking how learning-enabled systems should be evaluated, monitored, and governed. Questions that once belonged only to research communities are now business and policy questions. What evidence demonstrates readiness for deployment? How should system boundaries and operating limits be defined? What happens when systems encounter conditions not represented in training data? Who remains accountable when decisions are automated?

The panel discussion at SAE World Congress 2026 reflected this broader shift. Embodied AI is becoming an operational reality. The organizations that succeed will likely be those that combine technical innovation with disciplined approaches to safety, reliability, lifecycle management, and stakeholder trust.

For executives, this means embodied AI should be considered a strategic capability rather than an experimental novelty. For policymakers, it signals the need for frameworks that encourage innovation while protecting the public. For engineers and product leaders, it underscores the importance of designing systems that are not only intelligent, but dependable in the environments where they must perform.

\section{Perspectives from the Panel}
The SAE World Congress 2026 panel brought together participants with experience spanning mobility, robotics, artificial intelligence, safety engineering, systems integration, and standards development. Although their professional backgrounds differed, a consistent message emerged: embodied AI is advancing quickly, but successful deployment requires more than technical capability alone. It requires disciplined engineering, operational readiness, and public confidence.

From the automotive perspective, panelists noted that vehicles represent one of the most demanding environments for embodied AI. Roadways are open systems with constantly changing actors, weather conditions, infrastructure quality, and human behavior. Unlike controlled industrial settings, automotive systems must perform safely amid uncertainty and at scale. As a result, progress in automated driving, advanced driver assistance, and intelligent fleet operations depends on combining AI innovation with rigorous validation, clear operating boundaries, and resilient fallback strategies.

From the robotics perspective, panelists emphasized that many promising use cases already exist in logistics, manufacturing, healthcare, and field operations. In these environments, embodied AI can help machines adapt to spaces originally designed for humans rather than for fixed automation. Robots may need to navigate cluttered aisles, manipulate varied objects, coordinate with workers, or operate in changing conditions. This flexibility is one of the greatest strengths of embodied AI, but it also means systems must handle ambiguity, edge cases, and close-proximity interaction safely and consistently.

Panelists with AI and machine learning backgrounds stressed that model performance metrics alone are insufficient indicators of deployment readiness. A model may perform well in testing while still failing to generalize to rare scenarios, degraded sensors, novel environments, or unexpected user behavior. This reinforces the need to evaluate complete system performance rather than isolated algorithm benchmarks. Data quality, representativeness, monitoring, retraining controls, and lifecycle governance were identified as equally important considerations.

From a safety engineering standpoint, panelists highlighted the need to adapt proven methods to learning-enabled systems rather than discard them. Traditional approaches such as hazard analysis, requirements traceability, failure mode assessment, and layered mitigations remain highly relevant. However, they must be extended to account for data dependencies, probabilistic behavior, and performance drift over time. Several participants noted that the path forward is likely to involve integration between established safety disciplines and newer AI assurance methods.

The policy and standards perspective focused on trust, consistency, and accountability. Panelists observed that regulators and enterprise adopters increasingly seek practical evidence that embodied AI systems can be deployed responsibly. They are asking not only whether a system works under ideal conditions, but how it behaves under stress, how limitations are communicated, how updates are controlled, and who is accountable when issues arise. Standards development was viewed as an important mechanism for creating common expectations without unnecessarily slowing innovation.

Another recurring theme was the importance of cross-functional collaboration. Embodied AI cannot be delivered effectively when AI developers, hardware teams, safety specialists, operations leaders, and legal or policy stakeholders work in isolation. The panel noted that many deployment challenges emerge at the interfaces between disciplines rather than within any one discipline. Organizations that create shared language, integrated workflows, and common decision frameworks may be better positioned to scale adoption successfully.

Despite differing viewpoints, the panel’s overall perspective was optimistic. Embodied AI is expected to create meaningful benefits across transportation, industry, healthcare, and public services. The central question is no longer whether these systems will emerge, but how they can be engineered and governed in ways that are safe, trustworthy, and valuable.

\section{Key Deployment Challenges}
While the panel expressed optimism about the long-term potential of embodied AI, participants were equally clear that deployment at scale will require overcoming a set of practical and interconnected challenges. These challenges extend beyond algorithm design and touch engineering processes, operations, governance, and public acceptance.

\subsection{Safety Assurance in Dynamic Environments}
One of the most significant challenges is demonstrating safety in environments that are difficult to fully predict. Embodied AI systems must often operate in open or semi-structured settings where conditions change continuously. Vehicles encounter weather, construction zones, unusual traffic behavior, and vulnerable road users. Robots may face cluttered workspaces, shifting layouts, or close interaction with people. In these environments, it is not feasible to enumerate every possible scenario in advance.

Panelists noted that this changes the nature of assurance. Rather than proving perfect performance, organizations must establish confidence through layered evidence and structured assurance methods \cite{IntroductingMLFMEA, MLFMEAinAction}. This may include scenario testing, simulation, real-world operational data, redundancy, runtime monitoring, fallback strategies, and clearly defined operating limits.  Recent work on autonomous driving planners has shown the value of explicit hierarchical rule frameworks that prioritize safety-critical constraints while allowing lower-priority rules to be relaxed when conflicts arise \cite{Rulebooks}.

One challenge identified by panelists is that rigid rule-based regulatory assumptions may not fully reflect the realities of safe autonomous operation. In reality, safe driving depends on a hierarchy of behaviors in which safety considerations can, and often must, override strict legal compliance. Regulators should recognize that following the “rules of the road” is not always synonymous with ensuring safety; for example, both human drivers and automated systems may need to cross a double yellow line to avoid a hazard or momentarily exceed a low-speed constraint in a parking area to prevent becoming an obstruction. Accordingly, policies should move beyond rigid rule enforcement and instead require a transparent hierarchy of decision-making principles that clearly communicates how and when safety-driven trade-offs are made.

\subsection{Bridging the Gap Between Demonstration and Deployment}
Many embodied AI systems perform impressively in controlled demonstrations but struggle when introduced into live operations. Laboratory success does not always translate into robust field performance. Differences in lighting, sensor placement, user behavior, maintenance practices, infrastructure quality, or environmental variability can expose weaknesses that were not visible during development.

Deployment readiness requires solving 10 critical hardware challenges identified in the SAE J3329 report \cite{SAE_J3329_2026}, specifically balancing effectiveness (accuracy) with efficiency (throughput and power). Developers must also bridge the 'Sim-to-Real' gap by ensuring simulation fidelity matches hardware limitations; for example, a model trained on 1080p simulation will fail if the real-world camera is only 800x600. The use of Vision-Language-Action (VLA) models and transformers is recommended to translate high-level semantic intent into literal physical joint positions.

Panelists emphasized that deployment readiness should be treated as a separate milestone from prototype success. Achieving reliable real-world performance often requires iterative operational learning, stronger lifecycle controls, and disciplined change management.

\subsection{Data Quality and Generalization}

Because many embodied AI systems rely on machine learning, performance is strongly influenced by the quality and representativeness of training data. If relevant scenarios are missing, mislabeled, or underrepresented, systems may behave unpredictably when encountering those conditions in operation.

This challenge is especially important in safety-critical domains, where rare events may matter more than common ones. A vehicle that handles routine driving but fails in unusual merge scenarios, or a robot that functions normally but misses an infrequent human behavior pattern, may still present unacceptable risk. Panelists highlighted the need for deliberate data strategies that prioritize operational relevance rather than simply dataset scale.

In manufacturing, embodied AI provides essential flexibility for Small and Medium Enterprises (SMEs). Unlike rigid industrial robots, intelligent agents can be “taught” to quickly adapt to new tasks, allowing a single mobile robot to cover for missing staff and reduce downtime. To navigate data privacy constraints under strict regulations such as the European AI Act, organizations can adopt Private Federated Learning, which keeps the training and interpretation of sensitive data on local compute while aggregating only anonymized model weights for the global system, preserving both data richness and user privacy.\cite{learning-real-world-application}

\subsection{Human Trust and Interaction}

Embodied AI often operates in spaces shared with people. As a result, technical capability alone is insufficient. Systems must also behave in ways that humans can understand, anticipate, and trust.\cite{CanCarsGesture, TheRoadAhead, ComingIn,I_See_You} 

While industrial environments use buzzers and lights, robots in public spaces must utilize intuitive physical communication. Much like humans communicate through body language, autonomous vehicles signal intent through lane positioning and movement behavior. Standardizing these behaviors through a baseline of existing iconography (e.g., MIL-STD-2525) can significantly reduce training expenses and increase public trust.\cite{milstd2525e}

In mobility settings, this may include how a vehicle signals intent, responds to ambiguity, or behaves around pedestrians and cyclists. In robotics settings, it may include movement predictability, safe spacing, understandable interfaces, and confidence that the system will behave consistently. Trust can be built gradually through reliable performance, but it can be lost quickly through surprising or opaque behavior.

Trust in robots is built through physicality and body language. Rather than relying on external lights or buzzers, autonomous systems should use intuitive behaviors, such as positioning within a lane or movement style, to signal intent to pedestrians and other drivers.

\subsection{Standards, Regulation, and Accountability}

Another challenge identified by the panel is that governance frameworks are still evolving. Many standards and regulatory models were developed before learning-enabled physical systems became common. As a result, organizations may face uncertainty regarding acceptable evidence, reporting expectations, liability allocation, and lifecycle obligations after deployment.

Panelists viewed this as both a challenge and an opportunity. Clearer standards can accelerate adoption by reducing ambiguity and increasing confidence among customers, regulators, and the public.

Existing frameworks such as SAE J3016 ~\cite{SAE_J3016_2021} , ISO 21448~\cite{ISO_21448_2022}, ISO/PAS 8800~\cite{ISO_PAS_8800_2024}, ISO 13482~\cite{ISO_13482_2014}, SAE J3329~\cite{SAE_J3329_2026}, ISO/IEC TR 5469~\cite{ISO_IEC_TR_5469_2024}, and ISO/IEC 42001~\cite{ISO_IEC_42001_2023} provide useful foundations, though continued evolution is needed for embodied AI systems.

\subsection{Organizational Readiness}

Successful deployment also depends on internal capabilities. Many organizations still separate AI development, systems engineering, safety, operations, and compliance into independent functions. Embodied AI requires these groups to work together more closely than in traditional product models.

Without integrated decision-making, risks may emerge at the boundaries between teams. For example, a technically strong model may be deployed without sufficient operational monitoring, or a safe hardware platform may be paired with weak data governance. Panelists noted that organizational maturity may become as important as technical maturity in determining who succeeds.

\subsection{A Systems Challenge}

Across all of these topics, a common conclusion emerged: embodied AI deployment is fundamentally a systems challenge. Progress depends not only on smarter models, but on stronger engineering discipline, cross-functional alignment, and governance mechanisms that keep pace with innovation.

\subsection{Hardware Constraints}

Deployment requires a rigorous balance between effectiveness (the accuracy/precision of the AI model) and efficiency (throughput and power consumption). Unlike data-center AI, ground vehicles have strict power limits and must maintain high responsiveness with low latency.

\section{Emerging Consensus and Practical Recommendations}
Although panelists represented different sectors and technical backgrounds, several areas of broad consensus emerged during the discussion. These common themes provide a practical roadmap for organizations evaluating embodied AI initiatives today.

\subsection{Start with the Use Case, Not the Technology}

Panelists emphasized that successful programs begin with a clearly defined operational problem rather than with a desire to deploy AI for its own sake. Embodied AI should be selected where it offers meaningful value over conventional automation or software approaches. In some cases, traditional deterministic methods may remain the better choice. In others, learning-enabled systems may unlock flexibility, adaptability, or performance that would otherwise be difficult to achieve.

For leaders, this means beginning with mission needs, operating conditions, user expectations, and risk tolerance before selecting technical solutions.

\subsection{Define Operating Boundaries Early}

A recurring recommendation was the importance of explicitly defining where and under what conditions a system is expected to operate. Clear operating boundaries help shape data needs, validation strategies, user training, and safety controls.

For automotive systems, this may include road classes, weather conditions, speeds, and traffic contexts. For robotics systems, it may include workspace layouts, human proximity assumptions, object types, or environmental conditions. Systems that are well-bounded are generally easier to validate and govern than systems with undefined scope.

\subsection{Evaluate the Full Lifecycle}

Panelists consistently cautioned against viewing AI deployment as a one-time product release. Embodied AI systems require ongoing management across the lifecycle, including data updates, software changes, performance monitoring, maintenance, and incident response.

Organizations should establish ownership for post-deployment oversight early in the program rather than after launch. Continuous evaluation becomes especially important when models are retrained or environments evolve over time.

\subsection{Combine AI Innovation with Safety Discipline}

Another point of consensus was that organizations do not need to choose between innovation and rigor. Proven engineering methods such as hazard analysis, requirements management, redundancy strategies, and structured testing remain highly relevant. The opportunity lies in combining these methods with modern AI development practices.~\cite{IntroductingMLFMEA, MLFMEAinAction}

This integrated approach can help organizations move faster with greater confidence rather than slowing innovation.

\subsection{Prioritize Human Trust}

Whether in a vehicle cabin, warehouse aisle, or public setting, users and bystanders need confidence that embodied AI systems will behave predictably and responsibly. Trust is influenced by technical performance, but also by communication, transparency, and consistency.

Panelists encouraged organizations to include human interaction considerations early in design rather than treating them as cosmetic features added later.

\subsection{Build Cross-Functional Teams}

Embodied AI programs often fail when critical functions operate in isolation. Strong technical models can be undermined by weak operational processes, unclear accountability, or poor user integration.

The panel recommended cross-functional governance involving engineering, product, operations, safety, legal, and policy stakeholders. Shared decision frameworks can reduce friction and surface issues earlier.

\subsection{Engage with Standards Proactively}

Rather than waiting for regulation to mature fully, organizations were encouraged to participate in standards discussions, adopt emerging best practices early, and look to resources such as SAE J3321 for guidance on the unique Verification and Validation (V\&V) challenges introduced by learning-enabled systems. Doing so can improve internal readiness, reduce future compliance burdens, and help shape practical frameworks informed by operational realities.

\subsection{Move Deliberately, But Move}

Perhaps the clearest consensus message was that hesitation alone is not a strategy. Embodied AI is advancing across multiple industries, and waiting indefinitely may create competitive disadvantage. At the same time, reckless deployment can damage trust and slow progress.

The recommended path is deliberate advancement: pursue high-value use cases, apply disciplined engineering controls, learn through staged deployment, and build governance capabilities in parallel with technical capability.

\section{What Comes Next}
Embodied AI is entering a new phase. The question is shifting from whether intelligent machines can be built to how they can be deployed responsibly, scaled economically, and integrated into everyday operations. The technologies are advancing quickly, but long-term success will depend on the maturity of the surrounding ecosystem.  This includes decision architectures that can explain and justify actions under competing constraints, particularly in safety-critical mobility applications \cite{Rulebooks}.

In the near term, progress is likely to come through focused deployments in environments where value is clear and operating conditions can be reasonably bounded. In mobility, this may include specific driver assistance functions, structured freight operations, fleet optimization, or geofenced autonomy. In robotics, growth is expected in warehouses, manufacturing, inspection, healthcare support, and other use cases where repetitive tasks, labor constraints, or safety exposures create strong incentives for automation.

These narrower applications are important because they create operational learning. Organizations gain experience in validation methods, human factors, maintenance processes, data management, and governance. That experience can then support expansion into broader and more complex domains over time.

The panel also suggested that competitive advantage may increasingly come from execution rather than raw model capability. Access to advanced AI models is becoming more widespread, but the ability to integrate them safely into reliable products remains difficult. Companies that build strong systems engineering practices, disciplined deployment pipelines, and trust with customers and regulators may be better positioned than those focused solely on algorithmic novelty.

Standards and assurance methods are also expected to evolve significantly. As embodied AI deployments increase, regulators, insurers, customers, and industry groups will demand clearer evidence of safety and operational readiness. This is likely to accelerate the development of common terminology, evaluation frameworks, reporting expectations, and lifecycle controls. Organizations that engage early in these conversations may help shape practical standards while preparing themselves for future requirements.

Another area of future importance is workforce readiness. Embodied AI requires new combinations of skills that do not always exist within traditional organizational structures. Engineers may need greater familiarity with AI methods, while AI specialists may need stronger grounding in safety, controls, and systems thinking. Product leaders and policymakers will likewise need enough fluency to make informed decisions about opportunities and limitations. Talent models that bridge these disciplines may become a strategic differentiator.

The panel’s broader outlook was optimistic but measured. Embodied AI is unlikely to arrive as a single transformative event. Instead, it will advance through many incremental successes, lessons learned, and trust-building deployments. Some applications will mature quickly, while others will take longer due to technical complexity, economics, or regulatory constraints.

What appears most likely is that embodied AI will become increasingly normal. Intelligent vehicles, robots, and automated machines will gradually be integrated into sectors where they provide clear value, particularly where human interaction quality and trust are intentionally designed ~\cite{TheRoadAhead, I_See_You}.  The organizations that lead this transition will not necessarily be those with the most ambitious demonstrations, but those that combine innovation with reliability, safety discipline, and operational credibility.

For executives, policymakers, and technical leaders, the opportunity now is to shape that future deliberately. The decisions made today regarding standards, deployment models, workforce capability, and governance will influence how embodied AI is trusted and adopted in the years ahead.

\section{Panelists and Contributors}

This white paper reflects the themes and insights discussed during the SAE World Congress 2026 panel session Embodied AI in Action. The session brought together contributors from industry, consulting, startups, and academia with experience spanning automotive, robotics, artificial intelligence, machine learning operations, safety engineering, standards, and systems deployment.

\subsection{Panel Organizer and Moderator}
\paragraph{Panel organizers:}
\begin{itemize}
    \item \textbf{Jan-Mou Li--Independent}
    \item \textbf{Wei Tong--General Motors}
\end{itemize}

\paragraph{Moderator:} \textbf{Jan-Mou Li}

\subsection{Panelists}

\begin{itemize}
    \item \textbf{Paul Schmitt--Safety engineering, Reynolds and Moore}: safety of autonomous systems, machine learning assurance;
    \item \textbf{Wei Tong--Staff Researcher, General Motors}: 
applied AI and machine learning;
    \item \textbf{Majed Mohammed--Functional Safety and AI Safety leader, Aptiv}: Advanced automotive technology, automated driving systems;
    \item \textbf{Akshay Chalana--Co-founder, Saphira AI}: Artificial intelligence deployment, applied machine learning;
    \item \textbf{Arpan Kusari--Research Faculty, University of Michigan}: Autonomous vehicles, Physical AI, Sensing, Perception. 
\end{itemize}

\subsection{Invited Panelist Unable to Attend}

\textbf{Edward Griffor--National Institute of Standards and Technology (NIST)}: 
Invited participant who was unable to attend the live session

\subsection{Cross-Industry Perspective}

Collectively, the panel participants represent experience across a broad range of sectors including automotive, robotics, aerospace, industrial automation, energy, defense, research, and emerging AI ventures. This diversity of backgrounds contributed to a balanced discussion focused not only on technical capability, but also on real-world deployment, safety assurance, governance, and public trust.

\subsection{Shared Mission}

Despite differing domains and viewpoints, the panel was aligned around a common objective: advancing the safe, practical, and trustworthy adoption of embodied AI systems. The discussion emphasized that progress will depend on collaboration across disciplines that have historically operated separately, including AI development, systems engineering, safety, operations, and policy.



\bibliographystyle{ieeetr}
\bibliography{Embodied_AI_References}

@article{CanCarsGesture,
  author  = {Schmitt, Paul and Britten, Nicholas and Jeong, JiHyun and Coffey, Amelia and Clark, Kevin and Kothawade, Shweta Sunil and Grigore, Elena Corina and Khaw, Adam and Konopka, Christopher and Pham, Linh and Ryan, Kim and Schmitt, Christopher and Frazzoli, Emilio},
  title   = {Can Cars Gesture? A Case for Expressive Behavior Within Autonomous Vehicle and Pedestrian Interactions},
  journal = {IEEE Robotics and Automation Letters},
  year    = {2022},
  volume  = {7},
  number  = {2},
  pages   = {1416--1423},
  doi     = {10.1109/LRA.2021.3138161}
}

@inproceedings{ComingIn,
  author    = {Lee, Seong Hee and Britten, Nicholas and Block, Avram and Pandya, Aryaman and Jung, Malte F. and Schmitt, Paul},
  title     = {Coming In! Communicating Lane Change Intent in Autonomous Vehicles},
  booktitle = {Companion of the 2023 ACM/IEEE International Conference on Human-Robot Interaction},
  year      = {2023},
  pages     = {394--398},
  doi       = {10.1145/3568294.3580113}
}

@inproceedings{I_See_You,
author = {Block, Avram and Lee, Seonghee and Pandya, Aryaman and Schmitt, Paul},
title = {I See You! Design Factors for Supporting Pedestrian-AV Interaction at Crosswalks},
year = {2023},
isbn = {9781450399708},
publisher = {Association for Computing Machinery},
address = {New York, NY, USA},
url = {https://doi.org/10.1145/3568294.3580107},
doi = {10.1145/3568294.3580107},
booktitle = {Companion Proceedings of the 2023 ACM/IEEE International Conference on Human-Robot Interaction},
pages = {364–368},
numpages = {5},
location = {Stockholm, Sweden},
series = {HRI Companion '23}
}

@inproceedings{IntroductingMLFMEA,
  author    = {Schmitt, Paul and Seifert, Heinz Bodo and Lopez, Jerry and Bijelic, Mario and Heide, Felix and Pennar, Krzysztof},
  title     = {Introducing the ML FMEA},
  booktitle = {WCX SAE World Congress Experience},
  year      = {2025},
  month     = {April},
  publisher = {SAE International},
  doi       = {10.4271/2025-01-8078},
  url       = {https://doi.org/10.4271/2025-01-8078}
}

@techreport{MLFMEAinAction,
  author = {Schmitt, Philipp and Seifert, Hannes and Bijelic, Mario and others},
  title = {The ML FMEA in Action: Lessons from Applications of Machine Learning Safety},
  institution = {SAE International},
  year = {2026},
  month = {April},
  number = {2026-01-0079},
  doi = {10.4271/2026-01-0079},
  url = {https://saemobilus.sae.org/papers/ml-fmea-action-lessons-applications-machine-learning-safety-2026-01-0079}
}

@INPROCEEDINGS{TheRoadAhead,
  author={Block, Avram and Joshi, Swapna and Tabone, Wilbert and Pandya, Aryaman and Lee, Seonghee and Patil, Vaidehi and Britten, Nicholas and Schmitt, Paul},
  booktitle={2023 32nd IEEE International Conference on Robot and Human Interactive Communication (RO-MAN)}, 
  title={The Road Ahead: Advancing Interactions between Autonomous Vehicles, Pedestrians, and Other Road Users}, 
  year={2023},
  volume={},
  number={},
  pages={16-23},
  doi={10.1109/RO-MAN57019.2023.10309535}
}

@misc{SAE_J3016_2021,
  author       = {{SAE International}},
  title        = {{Taxonomy and Definitions for Terms Related to Driving Automation Systems for On-Road Motor Vehicles}},
  howpublished = {SAE Standard J3016\_202104},
  year         = {2021},
  note         = {Revised April 2021}
}

@misc{ISO_21448_2022,
  author       = {{International Organization for Standardization}},
  title        = {{Road Vehicles -- Safety of the Intended Functionality}},
  howpublished = {ISO 21448:2022},
  year         = {2022},
  note         = {SOTIF}
}

@misc{ISO_PAS_8800_2024,
  author       = {{International Organization for Standardization}},
  title        = {{Road Vehicles -- Safety and Artificial Intelligence}},
  howpublished = {ISO/PAS 8800:2024},
  year         = {2024}
}

@misc{ISO_13482_2014,
  author       = {{International Organization for Standardization}},
  title        = {{Robots and Robotic Devices -- Safety Requirements for Personal Care Robots}},
  howpublished = {ISO 13482:2014},
  year         = {2014}
}

@misc{ISO_IEC_42001_2023,
  author       = {{International Organization for Standardization and International Electrotechnical Commission}},
  title        = {{Information Technology -- Artificial Intelligence -- Management System}},
  howpublished = {ISO/IEC 42001:2023},
  year         = {2023}
}

@INPROCEEDINGS{Rulebooks,
  author={Hajieghrary, Hadi and Walter, Benedikt and Schmitt, Paul},
  booktitle={2026 IEEE/SICE International Symposium on System Integration (SII)}, 
  title={From Specification to Certification: TORQ-Ordered Rulebooks and Robust HOCBF Optimization for Safe Autonomous Driving}, 
  year={2026},
  volume={},
  number={},
  pages={881-886},
  doi={10.1109/SII64115.2026.11404471}
}

@misc{ISO_IEC_TR_5469_2024,
  author       = {{International Organization for Standardization and International Electrotechnical Commission}},
  title        = {{Artificial Intelligence -- Functional Safety and AI Systems}},
  howpublished = {ISO/IEC TR 5469:2024},
  year         = {2024},
  note         = {Edition 1, Published January 2024},
  institution  = {ISO/IEC JTC 1/SC 42}
}

@misc{SAE_J3329_2026,
  author       = {{SAE International}},
  title        = {{Artificial Intelligence in Ground Vehicles: Technical, Operational, and Regulatory Challenges}},
  howpublished = {SAE Standard J3329},
  year         = {2026},
  note         = {Projected publication date: May 2026; forthcoming}
}

@misc{milstd2525e,
  author = {Department of Defense},
  title = {MIL-STD-2525E, Interface Standard: Joint Military Symbology},
  year = {2022},
  month = {Dec},
  url = {https://quicksearch.dla.mil/qsDocDetails.aspx?ident_number=114934},
}

@misc{learning-real-world-application,
title = {Private Federated Learning In Real World Application – A Case Study},
author = {An Ji and Bortik Bandyopadhyay and Congzheng Song and Natarajan Krishnaswami and Prabal Vashisht and Rigel Smiroldo and Isabel Litton and Sayantan Mahinder and Mona Chitnis and Andrew W Hill},
year = {2025},
URL = {https://arxiv.org/abs/2502.04565}
}

@article{duan2022survey,
  title={A survey of embodied ai: From simulators to research tasks},
  author={Duan, Jiafei and Yu, Samson and Tan, Hui Li and Zhu, Hongyuan and Tan, Cheston},
  journal={IEEE Transactions on Emerging Topics in Computational Intelligence},
  volume={6},
  number={2},
  pages={230--244},
  year={2022},
  publisher={IEEE}
}

@article{chrisley2003embodied,
  title={Embodied artificial intelligence},
  author={Chrisley, Ron},
  journal={Artificial intelligence},
  volume={149},
  number={1},
  pages={131--150},
  year={2003},
  publisher={Elsevier}
}

@article{pfeifer2004embodied,
  title={Embodied artificial intelligence: Trends and challenges},
  author={Pfeifer, Rolf and Iida, Fumiya},
  journal={Lecture notes in computer science},
  pages={1--26},
  year={2004},
  publisher={Springer}
}

@article{bevly2016lane,
  title={Lane change and merge maneuvers for connected and automated vehicles: A survey},
  author={Bevly, David and Cao, Xiaolong and Gordon, Mikhail and Ozbilgin, Guchan and Kari, David and Nelson, Brently and Woodruff, Jonathan and Barth, Matthew and Murray, Chase and Kurt, Arda and others},
  journal={IEEE Transactions on Intelligent Vehicles},
  volume={1},
  number={1},
  pages={105--120},
  year={2016},
  publisher={IEEE}
}

@article{stryszowski2020framework,
  title={A framework for self-enforced interaction between connected vehicles: Intersection negotiation},
  author={Stryszowski, Marcin and Longo, Stefano and Velenis, Efstathios and Forostovsky, Gregory},
  journal={IEEE Transactions on Intelligent Transportation Systems},
  volume={22},
  number={11},
  pages={6716--6725},
  year={2020},
  publisher={IEEE}
}

@article{lackner2024review,
  title={Review of autonomous mobile robots in intralogistics: state-of-the-art, limitations and research gaps},
  author={Lackner, Thorge and Hermann, Julian and Kuhn, Christian and Palm, Daniel},
  journal={Procedia CIRP},
  volume={130},
  pages={930--935},
  year={2024},
  publisher={Elsevier}
}

@article{morais2025review,
  title={A review of robot fleet management},
  author={Morais, Paulo HC and Vivaldini, Kelen CT and Kato, Edilson RR and Inoue, Roberto S},
  journal={IEEE Access},
  year={2025},
  publisher={IEEE}
}

@article{bodenhagen2014adaptable,
  title={An adaptable robot vision system performing manipulation actions with flexible objects},
  author={Bodenhagen, Leon and Fugl, Andreas R and Jordt, Andreas and Willatzen, Morten and Andersen, Knud A and Olsen, Martin M and Koch, Reinhard and Petersen, Henrik G and Kr{\"u}ger, Norbert},
  journal={IEEE transactions on automation science and engineering},
  volume={11},
  number={3},
  pages={749--765},
  year={2014},
  publisher={IEEE}
}

@article{li2019survey,
  title={A survey of methods and strategies for high-precision robotic grasping and assembly tasks—Some new trends},
  author={Li, Rui and Qiao, Hong},
  journal={IEEE/ASME Transactions on Mechatronics},
  volume={24},
  number={6},
  pages={2718--2732},
  year={2019},
  publisher={IEEE}
}

@article{seo2018drone,
  title={Drone-enabled bridge inspection methodology and application},
  author={Seo, Junwon and Duque, Luis and Wacker, Jim},
  journal={Automation in construction},
  volume={94},
  pages={112--126},
  year={2018},
  publisher={Elsevier}
}

@article{lattanzi2017review,
  title={Review of robotic infrastructure inspection systems},
  author={Lattanzi, David and Miller, Gregory},
  journal={Journal of Infrastructure Systems},
  volume={23},
  number={3},
  pages={04017004},
  year={2017},
  publisher={American Society of Civil Engineers}
}

@article{aymerich2023socially,
  title={Socially assistive robots’ deployment in healthcare settings: a global perspective},
  author={Aymerich-Franch, Laura and Ferrer, Iliana},
  journal={International Journal of Humanoid Robotics},
  volume={20},
  number={01},
  pages={2350002},
  year={2023},
  publisher={World Scientific}
}

@article{holland2021service,
  title={Service robots in the healthcare sector},
  author={Holland, Jane and Kingston, Liz and McCarthy, Conor and Armstrong, Eddie and O’dwyer, Peter and Merz, Fionn and McConnell, Mark},
  journal={Robotics},
  volume={10},
  number={1},
  pages={47},
  year={2021},
  publisher={MDPI}
}

@article{ha2018recurrent,
  title={Recurrent world models facilitate policy evolution},
  author={Ha, David and Schmidhuber, J{\"u}rgen},
  journal={Advances in neural information processing systems},
  volume={31},
  year={2018}
}

\appendix
\section{Glossary of Terms}

This glossary is intended to help readers from business, policy, and non-technical backgrounds interpret terminology commonly used in discussions about embodied AI.

\paragraph{Artificial Intelligence (AI)}
A broad field of technology focused on creating systems that perform tasks typically associated with human intelligence, such as perception, reasoning, learning, planning, and decision-making.

\paragraph{Machine Learning (ML)}
A subset of AI in which systems learn patterns from data rather than relying only on explicitly programmed rules. Machine learning is commonly used for perception, prediction, and classification tasks.

\paragraph{Embodied AI}
AI systems that perceive, decide, and act through a physical machine operating in the real world. Examples include autonomous vehicles, mobile robots, drones, and intelligent industrial equipment.

\paragraph{Autonomous Vehicle}
A vehicle capable of performing some or all driving tasks using sensors, software, and control systems. Levels of automation may range from advanced driver assistance to highly automated operation in specific environments.

\paragraph{Robotics}
The design and operation of machines that can sense, move, and interact with the physical world. Robotics often combines mechanics, controls, sensing, and AI.

\paragraph{Perception System}
The portion of a system responsible for interpreting sensor inputs such as cameras, radar, lidar, microphones, or other data sources in order to understand the surrounding environment.

\paragraph{Planning System}
The component that determines what the machine should do next. In a vehicle, this may include selecting a path or maneuver. In a robot, it may include navigation or task sequencing.

\paragraph{Control System}
The component that converts decisions into physical action, such as steering, braking, acceleration, arm motion, or locomotion.

\paragraph{Sensor Fusion}
The combination of information from multiple sensors to improve understanding of the environment and increase robustness.

\paragraph{Operational Design Domain (ODD)}
The specific conditions under which a system is intended to operate safely. This may include road type, weather, lighting, speed range, geography, or facility constraints.

\paragraph{Validation}
The process of determining whether a system is suitable for its intended use in the real world.

\paragraph{Verification}
The process of determining whether a system was built according to its specified requirements and design.

\paragraph{Safety Assurance}
The structured body of evidence, analysis, testing, and process controls used to justify that a system can be deployed with acceptable risk.

\paragraph{Runtime Monitoring}
Mechanisms that observe system behavior during operation to detect faults, degraded performance, or conditions outside intended limits.

\paragraph{Fallback Strategy}
A predefined safe response used when the system encounters a fault, uncertainty, or scenario it cannot handle reliably.

\paragraph{Human Factors}
The study of how people interact with systems, including usability, trust, workload, communication, and predictable behavior.

\paragraph{Explainability}
The degree to which a system’s behavior or outputs can be understood by humans.

\paragraph{Generalization}
The ability of an AI system to perform effectively in new situations that differ from the data used during development.

\paragraph{Edge Case}
An unusual or uncommon scenario that may occur infrequently but can still be important for safety or reliability.

\paragraph{Governance}
The policies, responsibilities, oversight mechanisms, and decision processes used to manage technology responsibly throughout its lifecycle.

\paragraph{Lifecycle Management}
The ongoing management of a system after development, including updates, monitoring, maintenance, retraining, incident response, and retirement.

\paragraph{Standards}
Consensus-based documents that define common terminology, methods, expectations, or requirements to improve consistency, safety, and interoperability across an industry.

\end{document}